# 基于对偶流形重排序的无监督特征选择算法

梁云辉[1,2]　甘舰文[1,2]　陈　艳[3]　周　芃[4]　杜　亮[1,2]

1 山西大学计算机与信息技术学院　太原 030006
2 山西大学大数据科学与产业研究院　太原 030006
3 四川大学计算机学院　成都 610065
4 安徽大学计算机科学与技术学院　合肥 230601
(2350629530@qq.com)

**摘　要**　在许多数据分析任务中,经常会遇到高维数据。特征选择技术旨在从原始高维数据中找到最具代表性的特征,但由于缺乏类标签信息,相比有监督场景,在无监督学习场景中选择合适的特征困难得多。传统的无监督特征选择方法通常依据某些准则对样本的特征进行评分,在这个过程中样本是被无差别看待的。然而这样做并不能完全捕捉数据的内在结构,不同样本的重要性应该是有差异的,并且样本权重与特征权重之间存在一种对偶关系,它们会互相影响。为此,提出了一种基于对偶流形重排序的无监督特征选择算法(Unsupervised Feature Selection Algorithm based on Dual Manifold Re-Ranking,DMRR),分别构建不同的相似性矩阵来刻画样本与样本、特征与特征、样本与特征的流形结构,并结合样本与特征的初始得分进行流形上的重排序。将 DMRR 与 3 种原始无监督特征选择算法以及 2 种无监督特征选择后处理算法进行比较,实验结果表明样本重要性信息、样本与特征之间的对偶关系有助于实现更优的特征选择。

# Unsupervised Feature Selection Algorithm Based on Dual Manifold Re-ranking

LIANG Yunhui[1,2], GAN Jianwen[1,2], CHEN Yan[3], ZHOU Peng[4] and DU Liang[1,2]

1 College of Computer and Information Technology, Shanxi University, Taiyuan 030006, China
2 Institute of Big Data Science and Industry, Shanxi University, Taiyuan 030006, China
3 College of Computer, Sichuan University, Chendu 610065, China
4 College of Computer Science and Technology, Anhui University, Hefei 230601, China

**Abstract**　High dimensional data is often encountered in many data analysis tasks. Feature selection techniques aim to find the most representative features from the original high-dimensional data. Due to the lack of class label information, it is much more difficult to select suitable features in unsupervised learning scenarios than in supervised scenarios. Traditional unsupervised feature selection methods usually score the features of samples according to certain criteria in which samples are treated indiscriminately. However, these approaches cannot capture the internal structure of data completely. The importance of different samples should vary. There is a dual relationship between weight of sample and feature that will influence each other. Therefore, an unsupervised feature selection algorithm based on dual manifold re-ranking(DMRR) is proposed in this paper. Different similarity matrices are constructed to depict the manifold structures on samples and samples, features and features, and samples and features respectively. Then manifold re-ranking is carried out by combining the initial scores of samples and features. By comparing DMRR with three original unsupervised feature selection algorithms and two unsupervised feature selection post-processing algorithms, experimental results verify that importance information of different samples and the dual relationship between sample and feature are helpful to achieve better feature selection.

**Keywords**　Dual, Manifold learning, Re-ranking, Feature selection, Unsupervised learning

到稿日期:2022-10-18　返修日期:2022-11-12
基金项目:国家自然科学基金面上项目(61976129,62176001)
This work was supported by the National Natural Science Foundation of China(61976129,62176001).
通信作者:杜亮(duliang@sxu.edu.cn)

## 1 引言

近年来，随着信息技术的飞速发展，高维数据在图像处理、视频分析、基因表达数据分析、时间序列预测等领域大量出现。作为处理高维数据的有效工具，特征选择旨在从原始高维特征空间中找到一组信息含量丰富的特征子集，从而降低数据的维度并且提升算法的性能。根据标签的可用性，特征选择算法可分为有监督算法[1]、半监督算法[2]和无监督算法[3]。根据特征选择策略的不同，特征选择方法可以大致分为过滤式方法[4]、包装式方法[5]和嵌入式方法[6-7]。由于缺乏标签信息，无监督特征选择通常被认为比有监督特征选择更具有挑战性。本文主要关注无监督特征选择。

从数据角度来看，代表性的无监督特征选择算法可分为四大类[8]：基于相似性的方法[9-10]、基于信息理论的方法[11]、基于稀疏学习的方法[12]和基于统计的方法[13]。He 等[9]提出了一种基于拉普拉斯得分的特征选择算法（Laplacian Score for Feature Selection，LapScore）；Yao 等[10]提出了一种局部线性嵌入评分算法（LLEScore）；Lim 等[11]结合互信息提出了一种基于特征依赖的无监督特征选择算法（Feature Dependency-based Unsupervised Feature Selection，DUFS）；Cai 等[12]提出了一种多簇特征选择算法（MCFS）；He 等[13]提出了两种新的特征选择算法，分别为拉普拉斯正则化 A-最优特征选择（Laplacian Regula-rized A-Optimal Feature Selection，LapAOFS）和拉普拉斯正则化 D-最优特征选择（Laplacian Regularized D-Optimal Feature Selection，LapDOFS）。与上述直接对数据进行特征选择的算法不同，无监督特征选择算法中存在一类特别的算法——后处理算法，该算法可应用于现有的特征选择算法，以改进原始特征评分。Wang 等[14]提出了一种全局冗余最小化框架（Global Redundancy Minimization，GRM）用于特征选择；Nie 等[15]提出了一种基于全局冗余最小化的自动加权特征选择框架（Auto-weighted Feature Selection Framework via Global Redundancy Minimization，AGRM）。

尽管这些研究取得了巨大的进展，但是通过观察可以发现，无论是现有的无监督特征选择算法还是后处理算法都存在以下两点不足。1)现有方法在特征选择过程中仅区分特征单方面的重要性，不区分样本间的重要性，即所有样本都采用相同的权重进行处理，这样做既无法突出高质量样本的贡献，也无法降低噪声[1]和异常样本的影响，会限制学习算法的性能。实际应用中数据样本重要性相同这一假设往往并不成立。以密度聚类[16]为例，距离中心点较近的样本被赋予较高的权重；以主动学习[17-18]为例，少量代表性样本被选择标注并用于后续的高质量学习过程。2)现有方法忽略了样本与特征之间普遍存在对偶关系，无法利用样本层的结构信息来进一步指导特征的选择以致于得不到理想的特征子集。现有研究表明联合聚类[19-20]在聚类过程中通过结合特征样本对偶关系可以有效提高样本侧聚类效果。

针对上述两个问题，本文提出了一种基于对偶流形重排序的无监督特征选择算法，不仅考虑了样本、特征的重要性信息，还考虑了样本与特征之间的对偶关系。对于普遍存在的异常点、噪声等数据质量问题，引入了样本权重来区分不同样本的重要性，以减轻低质量样本的干扰进而更加精准地刻画数据的内在结构。同时，我们借助样本与特征间的对偶关系，一方面通过选择重要特征来提高样本侧簇结构刻画能力与对应的聚类结果，另一方面借助更好的聚类结果进一步指导特征选择过程进而获得更佳的特征子集。具体来讲，受对偶学习的启发，我们分别构造样本层、特征层和样本与特征层相似性图来捕捉数据的内在结构。值得一提的是，与现有方法构造 $n\times n$ 的图不同，本文在构建样本与特征层相似性图时，设计了一种 $n\times d$ 的二部图来描述其结构，可以更细致地刻画样本与特征之间的对偶关系。在得到这些图后利用流形学习分别捕捉其内在结构，并结合原始样本得分与特征得分对特征进行重排序。该方法作为一种后处理算法，可以对特征选择算法得到的特征评分进行重排序以实现聚类性能的提升。在多个数据集上进行了实验，结果表明样本权重信息、样本与特征之间的对偶关系对于特征选择任务是非常重要的。

本文的主要贡献包括 3 个方面：

(1)现有方法一方面没有考虑不同样本应具有不同的重要性，另一方面没有考虑样本空间与特征空间的对偶关系。我们认为不同样本的重要性是有差别的，并且样本与特征之间存在某种对偶关系。这两种信息有助于更好地描述数据的内在结构。

(2)设计了一种基于对偶流形重排序的无监督特征选择算法，分别构造样本层相似图、特征层相似图、样本与特征之间二部图，并利用其流形结构结合样本与特征初始得分来进行特征的流形重排序。

(3)在多个公开数据集上将本文方法与多种对比方法进行比较，结果不仅表明了所提方法的有效性，而且验证了样本重要性信息、样本与特征之间对偶关系是有助于特征选择任务的。此外本文所提出的算法源码已公开[1]。

## 2 相关工作

### 2.1 无监督特征选择方法

为了处理高维数据，研究者们提出了许多无监督特征选择算法。Luo 等[21]提出通过重构非负权重图来刻画每个邻域的内在几何结构，并对其拉普拉斯矩阵施加秩约束以实现理想的邻域分配。为了避免稀疏正则化引入的额外超参数，Li 等[22]设计了一个特征滤波器来估计图像特征的权重，并通过理论分析表明稀疏正则化可由特征滤波器变换得到。Li 等[23]考虑到样本分布和以更有效顺序使用样本训练学习方法的潜在效果，结合自步学习与子空间学习来进行特征选择。Wahid 等[24]提出了一种具有鲁棒数据重建的无监督特征选择算法（Unsupervised Feature Selection with Robust Data Reconstruction，UFS-RDR），它可以最小化图正则化加权数据重建误差函数，其使用 Huber 型权重函数建模，该权重函数会降低距离较大的聚类观测值的权重。该方法在数据存在异常

[1] https://gitee.com/lyhspace/dmrr

值的情况下能取得较好的效果。Xie 等[25]提出一种无监督特征选择算法 SCFS(Standard Deviation and Cosine Similarity Based Feature Selection),SCFS 使用标准差获得不同类之间的区别,使用余弦相似性表示特征之间的冗余。Bao 等[26]结合邻域信息和标记相关性提出了在线多标记特征选择选择算法,能够实现在线评价动态候选特征。Lv 等[27]提出了一种基于邻域区间扰动融合的无监督特征选择算法框架,该模型可实现特征的最终得分与近似数据区间的联合学习。Han 等[28]结合自动编码器和组拉索回归任务提出了一种被称为自动编码器特征选择器的无监督特征选择算法,该方法可通过挖掘特征中的线性和非线性信息来选择最重要的特征。Beiranvand 等[29]提出了一种基于主成分分析的无监督特征选择算法(Unsupervised Feature Selection Using Principal Component Analysis, UFSPCA),该方法使用 PCA(Principal Component Analysis)创建不相关和正交特征,根据原始特征与不相关特征的相似度构建加权二部图,并使用匈牙利算法获得具有最大权重的匹配来选定特征。Zhao 等[30]认为利用原始特征构建的图来学习子空间存在一定的局限性,为此提出了一种自适应图学习策略来学习数据结构信息更准确的高质量图,并增加不相关约束来增强模型的可判别性。Miao 等[31]提出了一种图正则化局部线性嵌入(Graph Regularized Local Linear Embedding, GLLE)方法,该方法将局部线性嵌入和特征子空间中的流形正则化集成到一个统一的框架中,通过加强无监督特征选择与特征子空间之间的关系来更有效地选择相关特征。

### 2.2 特征重排序方法

无监督特征选择算法中存在的一类方法为后处理方法。Wang 等[14]提出了一种无监督全局和局部判别特征选择方法,他们认为所有的特征都能基于分数进行排序,并设计了一种新的全局冗余最小化算法(Global Redundancy Minimization, GRM),旨在集成任何类型的特征排序分数,通过最小化指定特征排序的全局特征冗余、最大化原始分数与指定分数的一致性来提升原特征选择算法的性能。Nie 等[15]提出了一种基于全局冗余最小化的自动化加权特征选择算法(A Novel Auto-weighted Feature Selection Framework via Global Redundancy Minimization, AGRM),与其他特征选择方法不同,AGRM 算法可以真正选择具有代表性的非冗余特征,从全局角度来看,可以大大减少特征之间的冗余。这两种算法都是后处理算法,可以对其他特征选择算法输出的特征得分进行重排序,从而得到更好的聚类性能。

可以看出,为了提升特征选择算法的性能,研究者们努力地进行各种尝试,但是现有的无监督特征选择算法以及后处理算法在依据各种评估准则、学习算法来选取特征时忽视了一些信息。

(1)不同样本的重要性。之前,学者们使用等权重样本来构造模型,这其实是有问题的。无论从聚类的角度还是分类的角度来看,样本的权重都是有差异的。以 K-MEANS 算法[32]为例,有的样本距离聚类中心点较近,而有的样本距离聚类中心点较远。因为聚类中心点是由样本构成的,所以哪个样本离得近,它的权重就会越高,反之,它的权重就会越低,这说明样本的重要性是有差别的。

(2)样本与特征之间存在对偶关系,即特征聚类结果和样本聚类结果相互依赖。当样本权重较大时,该样本的特征也应该具有较高的权重,反之亦然。考虑到样本与特征之间的对偶关系与样本权重的重要性,我们在构建模型时同时利用样本权重与特征权重信息,并结合流形排序提出了一种新的基于对偶流形重排序的无监督特征选择算法(DMRR)。DMRR 作为一种后处理方法,可以对其他算法得到的特征得分进行重排序,并且可以在一定程度上实现聚类性能的提升。

## 3 研究方法

### 3.1 基本定义

首先介绍一些在本文中使用的概念。矩阵用大写字母加粗表示,矢量用小写字母加粗表示。对于数据集 $\mathbf{X}\in R^{n\times d}$,$n$ 表示样本的数量,$d$ 表示特征的维度。数据中第 $i$ 个样本的第 $j$ 个特征可以表示为$\mathbf{X}_{ij}$,$\mathbf{X}$ 的第 $i$ 行表示为$\mathbf{X}_{i.}\in R^{1\times d}$,$\mathbf{X}$ 的第 $j$ 列表示为$\mathbf{X}_{.j}\in R^{n\times 1}$。$\mathbf{u}\in R^{n\times 1}$ 为样本得分向量,$\mathbf{v}\in R^{d\times 1}$ 为特征得分向量。$l_2$ 范数的定义为 $\|\mathbf{X}\|_2=\sqrt{\sum_{i=1}^{n}\sum_{j=1}^{n}X_{ij}^{2}}$。

### 3.2 基于对偶流行重排序的无监督特征选择算法

本文具体算法过程示意图如图 1 所示。

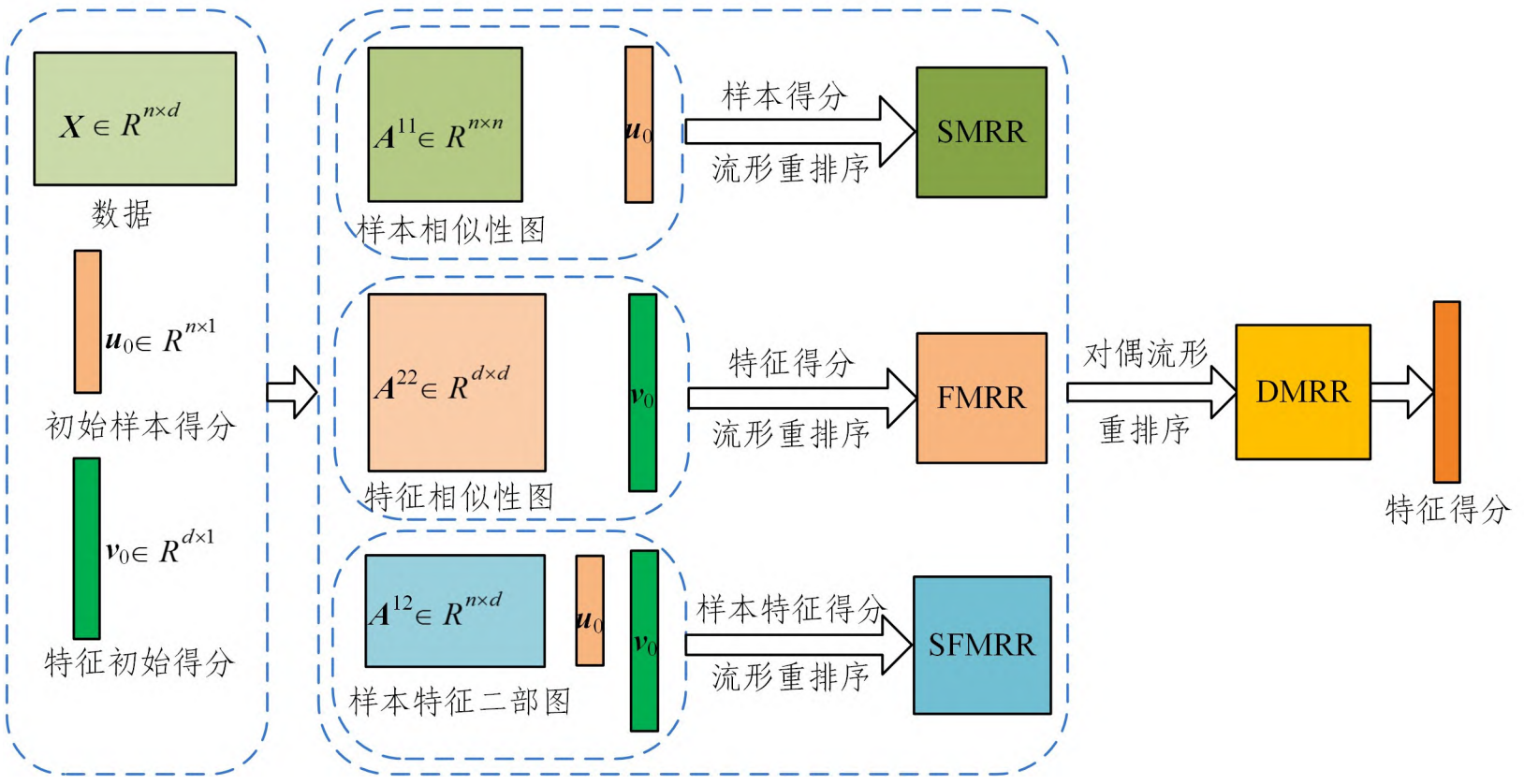

图 1 算法过程示意图

Fig. 1 Schematic diagram of algorithm process

3.2.1 样本流形重排序

传统的特征选择算法通常依据某种准则直接对数据的特征进行排序，而忽略了样本权重所富含的信息。我们认为样本的权重是有差异的，有的样本容易学习，有的样本不容易学习，并且一些研究也证明了这一点。那么在构建模型时，就应该考虑到样本的权重信息，因此，我们考虑结合流形排序来探索样本的内在流形结构并为样本计算权重。首先使用 $k$ 最近邻方法来构造样本相似性图 $\boldsymbol{A}^{11}$，$\boldsymbol{A}^{11}$ 中每个元素的定义如下：

$$\boldsymbol{A}_{ij}^{11}=\begin{cases}\mathrm{e}^{-\frac{\gamma\|\boldsymbol{X}_{i\cdot}-\boldsymbol{X}_{j\cdot}\|^2}{\delta^2}}, & \text{如果}\boldsymbol{X}_{i\cdot}\text{与}\boldsymbol{X}_{j\cdot}\text{是邻居}\\ 0, & \text{其他}\end{cases} \tag{1}$$

其中，$\gamma$ 为超参数，用于调节高斯函数的平滑程度；$\delta$ 为高斯函数的带宽，其值为所有样本两两欧氏距离的中值。然后根据 $\boldsymbol{A}^{11}$ 通过式(2)来刻画样本的流形结构。

$$\begin{aligned}&\min\sum_{i=1}^{n}\sum_{j=1}^{n}\boldsymbol{A}_{ij}^{11}\left(\frac{\boldsymbol{u}_i}{\sqrt{\boldsymbol{d}_i}}-\frac{\boldsymbol{u}_j}{\sqrt{\boldsymbol{d}_j}}\right)^2\\&\text{s. t. }\ \boldsymbol{u}^{\mathrm{T}}\ 1=1,\boldsymbol{u}\geqslant 0\end{aligned} \tag{2}$$

在此基础上联合初始样本得分向量 $\boldsymbol{u}_0$ 可以进一步得到样本流形重排序的目标函数 SMRR。

$$\begin{aligned}&\min\ \boldsymbol{u}^{\mathrm{T}}\boldsymbol{L}_u\boldsymbol{u}+\lambda\|\boldsymbol{u}-\boldsymbol{u}_0\|^2\\&\text{s. t. }\ \boldsymbol{u}^{\mathrm{T}}\ 1=1,\boldsymbol{u}\geqslant 0\end{aligned} \tag{3}$$

其中，$\boldsymbol{L}_u=\boldsymbol{I}-\boldsymbol{D}_u^{-\frac{1}{2}}\boldsymbol{A}^{11}\boldsymbol{D}_u^{-\frac{1}{2}}$ 是根据样本相似性图 $\boldsymbol{A}^{11}$ 得到的拉普拉斯矩阵；$\boldsymbol{D}_u$ 为 $\boldsymbol{u}^{11}$ 的度矩阵，其第 $i$ 个对角元素为 $\boldsymbol{D}_{u_{ii}}=\sum_{j=1}^{n}\boldsymbol{A}_{ij}^{11}$。第二项是一个正则项，用于根据样本得分先验信息 $\boldsymbol{u}_0$ 学习最终的样本得分向量 $\boldsymbol{u}$，$\lambda$ 为平衡参数。

3.2.2 特征流形重排序

针对特征之间的结构，本文使用 $k$ 最近邻方法来构建特征层相似性图 $\boldsymbol{A}^{22}$，其中每个元素的定义如下：

$$\boldsymbol{A}_{ij}^{22}=\begin{cases}\mathrm{e}^{-\frac{\gamma\|\boldsymbol{X}_{\cdot i}-\boldsymbol{X}_{\cdot j}\|^2}{\delta^2}}, & \text{如果}\boldsymbol{X}_{\cdot i}\text{与}\boldsymbol{X}_{\cdot j}\text{是邻居}\\ 0, & \text{其他}\end{cases} \tag{4}$$

其中，$\gamma$ 为超参数；$\delta$ 为高斯函数的带宽，其值为所有特征两两欧氏距离的中值。然后根据特征层相似性图 $\boldsymbol{A}^{22}$ 结合流形学习来捕捉特征之间的局部结构。

$$\begin{aligned}&\min\sum_{i=1}^{d}\sum_{j=1}^{d}\boldsymbol{A}_{ij}^{22}\left(\frac{\boldsymbol{v}_i}{\sqrt{\boldsymbol{d}_i}}-\frac{\boldsymbol{v}_j}{\sqrt{\boldsymbol{d}_j}}\right)^2\\&\text{s. t. }\ \boldsymbol{v}^{\mathrm{T}}\ 1=1,\boldsymbol{v}\geqslant 0\end{aligned} \tag{5}$$

通过变换，式(5)可进一步转化为：

$$\begin{aligned}&\min\ \boldsymbol{v}^{\mathrm{T}}\boldsymbol{v}-\boldsymbol{v}^{\mathrm{T}}\boldsymbol{D}_v^{-\frac{1}{2}}\boldsymbol{A}^{22}\boldsymbol{D}_v^{-\frac{1}{2}}\boldsymbol{v}\\&\text{s. t. }\boldsymbol{v}^{\mathrm{T}}\ 1=1,\boldsymbol{v}\geqslant 0\end{aligned} \tag{6}$$

其中，$\boldsymbol{D}_{v_{ii}}=\sum_{j=1}^{d}\boldsymbol{A}_{ij}^{22}$。然后结合特征之间的流形结构与初始特征得分向量 $\boldsymbol{v}_0$ 建立特征流形重排序的目标函数 FMRR。

$$\begin{aligned}&\min\ \boldsymbol{v}^{\mathrm{T}}\boldsymbol{L}_v\boldsymbol{v}+\lambda\|\boldsymbol{v}-\boldsymbol{v}_0\|^2\\&\text{s. t. }\ \boldsymbol{v}^{\mathrm{T}}\ 1=1,\boldsymbol{v}\geqslant 0\end{aligned} \tag{7}$$

其中，$\boldsymbol{L}_v=\boldsymbol{I}-\boldsymbol{D}_v^{-\frac{1}{2}}\boldsymbol{A}^{22}\boldsymbol{D}_v^{-\frac{1}{2}}$ 是根据特征相似性图 $\boldsymbol{A}^{22}$ 构建的拉普拉斯矩阵；$\boldsymbol{v}_0$ 是初始特征得分向量，$\boldsymbol{v}$ 是目标函数要学习的特征得分向量。

此外 Wang 等[14] 提出了一种特征选择框架(Global Redundancy Minimization，GRM)，其目标函数为：

$$\begin{aligned}&\min\ \frac{\boldsymbol{z}^{\mathrm{T}}\boldsymbol{A}\boldsymbol{z}}{\boldsymbol{z}^{\mathrm{T}}\boldsymbol{s}}\\&\text{s. t. }\boldsymbol{z}^{\mathrm{T}}\ 1=1,\boldsymbol{z}\geqslant 0\end{aligned} \tag{8}$$

其中，$\boldsymbol{s}\in R^{d\times 1}$ 是根据某种特征排序准则生成的初始特征得分向量，$\boldsymbol{z}\in R^{d\times 1}$ 是通过该模型计算得出的特征得分向量，$\boldsymbol{A}\in R^{d\times d}$ 为相似性矩阵。

从式(6)和式(8)可以看出这两种方法在思路上的不同，GRM 通过最大化输入特征得分与所求特征得分之间的一致性，同时最小化全局特征冗余信息来获得更好的特征得分。而我们认为一小部分的特征可以包含大量有用的聚类信息，应该为这一部分特征赋予较大的权重，给其他特征分配较小的权重。特征流形重排序目标函数的建立正是基于这一点。

3.2.3 基于二部图的样本特征流形重排序

我们利用二部图结构来描述样本与特征之间的关系。如图 2 所示，二部图是一种普遍存在的数据结构，可以用来表示两种实体类型之间的关系。在建模时，不仅应考虑同一类型实体之间的关系，还应该考虑不同类型实体之间的关系，这种关系会自然形成一个二部图，其中包含丰富的信息可供挖掘。

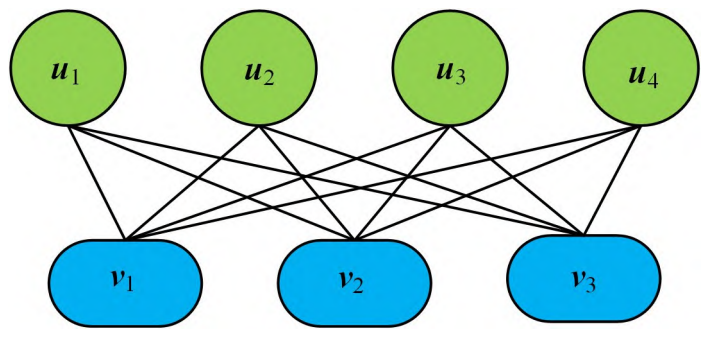

图 2 样本与特征之间二部图结构

Fig. 2 Structure of bipartite graph between sample and feature

每个特征的得分都与样本息息相关，针对这种结构，本文通过高斯函数来刻画样本与特征之间关系。高斯函数被广泛应用于机器学习任务中以捕捉数据之间的相似性，其简单高效且能使得样本与特征之间相似性呈高斯分布，即为少量相距较近的样本特征对分配较高的相似度而为大量相距较远的样本特征对分配较小的相似度，以实现对样本与特征之间关系更加细致的描述。首先构造样本与特征之间的二部图 $\boldsymbol{A}_{ij}^{12}$，其中每个元素的具体定义为：

$$\boldsymbol{A}_{ij}^{12}=\mathrm{e}^{-\frac{\gamma\|\boldsymbol{X}_{ij}-\bar{\boldsymbol{u}}_j\|^2}{\delta^2}} \tag{9}$$

其中，$\gamma$ 为高斯函数中的参数；$\bar{\boldsymbol{u}}_j=\frac{1}{n}\sum_{i=1}^{n}\boldsymbol{X}_{ij}$ 表示数据集中第 $j$ 个特征的平均值；$\delta$ 为高斯函数中的带宽，其值分 3 步确定：1)计算每个元素与该维特征均值的差值；2)计算每个特征维度差值的中值；3)取 $d$ 个中值的最大值即为 $\delta$。根据样本与特征的二部图可以捕捉样本与特征之间的流形结构。

$$\begin{aligned}&\min\sum_{i=1}^{n}\sum_{j=1}^{d}\boldsymbol{A}_{ij}^{12}\left(\frac{\boldsymbol{u}_i}{\sqrt{\boldsymbol{d}_i}}-\frac{\boldsymbol{v}_j}{\sqrt{\boldsymbol{d}_j}}\right)^2\\&\text{s. t. }\ \boldsymbol{u}^{\mathrm{T}}\ 1=1,\boldsymbol{u}\geqslant 0,\boldsymbol{v}^{\mathrm{T}}\ 1=1,\boldsymbol{v}\geqslant 0\end{aligned} \tag{10}$$

设计目标函数 SFMRR 利用样本与特征的对偶信息对 $\boldsymbol{u}$ 和 $\boldsymbol{v}$ 进行排序。具体形式如下：

$$\begin{aligned}&\min_{\boldsymbol{u},\boldsymbol{v}}\boldsymbol{u}^{\mathrm{T}}\boldsymbol{u}-2\boldsymbol{u}^{\mathrm{T}}\hat{\boldsymbol{D}}_u^{-\frac{1}{2}}\boldsymbol{A}^{12}\hat{\boldsymbol{D}}_v^{-\frac{1}{2}}\boldsymbol{v}+\boldsymbol{v}^{\mathrm{T}}\boldsymbol{u}+\lambda\|\boldsymbol{u}-\boldsymbol{u}_0\|^2+\\&\lambda\|\boldsymbol{v}-\boldsymbol{v}_0\|^2\\&\text{s. t. }\boldsymbol{u}^{\mathrm{T}}\ 1=1,\boldsymbol{u}\geqslant 0,\boldsymbol{v}^{\mathrm{T}}\ 1=1,\boldsymbol{u}\geqslant 0\end{aligned} \tag{11}$$

其中，$\hat{D}_{u_{ii}}=\sum_{j=1}^{d}A_{ij}^{12}$，$\hat{D}_{v_{ii}}=\sum_{i=1}^{n}A_{ij}^{12}$。此外，本模型还包括特征与样本之间的二部图，用样本与特征之间二部图的转置来表示。

3.2.4 对偶流形重排序

针对样本与特征之间存在的这种对偶关系，以及样本与样本之间、特征与特征之间、样本与特征之间的结构，本文结合样本得分流形重排序、特征得分流形重排序、样本特征得分流形重排序提出一种新的模型——基于对偶流形重排序的无监督特征选择算法(DMRR)。DMRR 作为一种后处理方法，可以对其他特征选择算法得到的特征得分进行改进，得到最终的特征得分 $v\in R^{d\times 1}$，从而实现对原始得分的重排序以提高算法的聚类结果。

DMRR 模型的具体形式如下：

$$\min_{u,v}\begin{bmatrix}u\\v\end{bmatrix}^{\mathrm{T}}L\begin{bmatrix}u\\v\end{bmatrix}+\lambda_1\left\|\begin{bmatrix}u\\v\end{bmatrix}-\begin{bmatrix}u_0\\v_0\end{bmatrix}\right\|^2 \quad (12)$$
$$\text{s.t. } u^{\mathrm{T}}1=1,u\geqslant 0,v^{\mathrm{T}}1=1,v\geqslant 0$$

模型中：$u\in R^{n\times 1}$ 表示样本原始得分经过 DMRR 模型运算后得到的样本得分，初始化为 $u=\frac{1}{n}$；$v\in R^{d\times 1}$ 作为模型的输出，是模型联合对偶学习和流形排序得出的特征评分，初始化为 $v=\frac{1}{d}$；$v_0\in R^{d\times 1}$ 作为模型的输入，是通过其他特征选择算法得到的初始特征评分；$u_0\in R^{n\times 1}$ 表示输入样本的初始得分，其值由 Li 等[33]提出的一种简单且独立于初始化的样本得分策略来确定。

该策略可简单描述为：1)对于每个维度特征中存在的负值，令其减去该维度特征中的最小值，从而将原始特征转换为非负值，并对每个维度的特征求和得到样本得分 $s\in R^{n\times 1}$；2)通过除以最大样本得分来正则化样本得分，即 $s_i=\frac{s_i}{\max(s)}$；3)通过 $u_i=s_i(1-s_i)$ 更新样本得分，得到最终样本得分 $u_0\in R^{n\times 1}$。

式(12)中 $L$ 表示由样本相似性图、特征相似性图、样本特征二部图得到的拉普拉斯矩阵。

$$L=\left[2I-\lambda_2\begin{bmatrix}D_u^{-\frac{1}{2}}A^{11}D_u^{-\frac{1}{2}} & \hat{D}_u^{-\frac{1}{2}}A^{12}\hat{D}_v^{-\frac{1}{2}}\\ \hat{D}_v^{-\frac{1}{2}}A^{21}\hat{D}_u^{-\frac{1}{2}} & D_v^{-\frac{1}{2}}A^{22}D_v^{-\frac{1}{2}}\end{bmatrix}\right] \quad (13)$$

模型中有两个平衡参数 $\lambda_1$ 和 $\lambda_2$，其中 $\lambda_1$ 用来控制样本特征得分的先验信息与最终得分的逼近程度。当 $\lambda_1$ 的值增大时，两者之间的差异将缩小，而 $\lambda_2$ 作用于不同相似性矩阵，其值的变化将影响算法的最大相似性保持。

## 3.3 模型的求解

模型的求解采用交替变量优化的方式，将拉普拉斯矩阵 $L$ 具体形式代入式(12)可以得到：

$$\min_{u,v}\begin{bmatrix}u\\v\end{bmatrix}^{\mathrm{T}}\left[2I-\lambda_2\begin{bmatrix}D_u^{-\frac{1}{2}}A^{11}D_u^{-\frac{1}{2}} & \hat{D}_u^{-\frac{1}{2}}A^{12}\hat{D}_v^{-\frac{1}{2}}\\ \hat{D}_v^{-\frac{1}{2}}A^{21}\hat{D}_u^{-\frac{1}{2}} & D_v^{-\frac{1}{2}}A^{22}D_v^{-\frac{1}{2}}\end{bmatrix}\right]\begin{bmatrix}u\\v\end{bmatrix}+\lambda_1\left\|\begin{bmatrix}u\\v\end{bmatrix}-\begin{bmatrix}u_0\\v_0\end{bmatrix}\right\|^2 \quad (14)$$
$$\text{s.t. } u^{\mathrm{T}}1=1,u\geqslant 0,v^{\mathrm{T}}1=1,v\geqslant 0$$

(1)固定 $v$，更新 $u$

式(14)可等价于求解以下问题：

$$\min_u u^{\mathrm{T}}[I-\lambda_2 D_u^{-\frac{1}{2}}A^{11}D_u^{-\frac{1}{2}}]u+u^{\mathrm{T}}u-\lambda_2 u^{\mathrm{T}}\hat{D}_u^{-\frac{1}{2}}A^{12}\hat{D}_v^{-\frac{1}{2}}v+v^{\mathrm{T}}v-\lambda_2 v^{\mathrm{T}}\hat{D}_v^{-\frac{1}{2}}A^{21}\hat{D}_u^{-\frac{1}{2}}u+\lambda_1\|u-u_0\|^2$$
$$\text{s.t. } u^{\mathrm{T}}1=1,u\geqslant 0 \quad (15)$$

其中，$A^{12}=A^{21\mathrm{T}}$，该问题可以进一步转换为如下问题：

$$\min_u u^{\mathrm{T}}A_u u+u^{\mathrm{T}}B_u \quad (16)$$
$$\text{s.t. } u^{\mathrm{T}}1=1,u\geqslant 0$$

其中，

$$A_u=2I+\lambda_1 I-\lambda_2 D_u^{-\frac{1}{2}}A^{11}D_u^{-\frac{1}{2}}$$
$$B_u=-2\lambda_2\hat{D}_u^{-\frac{1}{2}}A^{12}\hat{D}_v^{-\frac{1}{2}}v-2\lambda_1 u_0 \quad (17)$$

显然，式(16)是一个带线性约束的凸二次规划问题，该问题可使用现有的优化工具解决。

(2)固定 $u$，更新 $v$

由式(14)可得到以下优化问题：

$$\min_v v^{\mathrm{T}}[I-\lambda_2 D_v^{-\frac{1}{2}}A^{22}D_v^{-\frac{1}{2}}]v+v^{\mathrm{T}}v-\lambda_2 v^{\mathrm{T}}\hat{D}_v^{-\frac{1}{2}}A^{21}\hat{D}_u^{-\frac{1}{2}}u+u^{\mathrm{T}}u-\lambda_2 u^{\mathrm{T}}\hat{D}_u^{-\frac{1}{2}}A^{12}\hat{D}_v^{-\frac{1}{2}}v+\lambda_1\|v-v_0\|^2$$
$$\text{s.t. } v^{\mathrm{T}}1=1,v\geqslant 0 \quad (18)$$

在本步骤中使用的 $u$ 为前一步更新后的 $u$，式(18)可以等价转换为如下问题：

$$\min_u v^{\mathrm{T}}A_v v+v^{\mathrm{T}}B_v \quad (19)$$
$$\text{s.t. } v^{\mathrm{T}}1=1,u\geqslant 0$$

其中，

$$A_v=2I+\lambda_1 I-\lambda_2 D_v^{-\frac{1}{2}}A^{22}D_v^{-\frac{1}{2}}$$
$$B_v=-2\lambda_2\hat{D}_v^{-\frac{1}{2}}A^{21}\hat{D}_u^{-\frac{1}{2}}u-2\lambda_1 v_0 \quad (20)$$

与式(16)类似，此问题同样可以用现有的优化工具解决。

综上所述，DMRR 模型的求解流程如算法 1 所示。

**算法 1 基于对偶流形重排序的无监督特征选择算法**

输入：数据集 **X**，数据标签 **Y**，选择特征数量 m，数据特征得分 $v_0$，平衡参数 $\lambda_1$，$\lambda_2$

输出：得分高的前 m 个特征

1. 初始化 $u$，$v$，$u_0$；
2. 构造样本相似性图 $A^{11}$；
3. 构造特征相似性图 $A^{22}$；
4. 构造样本特征二部图 $A^{12}$；
5. 分别求解每个相似性图对应的度矩阵 $D_u$，$D_v$，$\hat{D}_u$，$\hat{D}_v$；
6. while not do convergence：
7. 　通过式(16)更新 $u$；
8. 　通过式(19)更新 $v$；
9. 　判断收敛 $\left|\frac{F(t-1)-F(t)}{F(t-1)}\right|<10^{-6}$，F 为式(12)目标函数；
10. end while
11. 对特征得分 $v$ 进行降序排序，选择前 m 个特征。

## 4 实验结果分析

### 4.1 数据集

为了衡量本文提出的基于对偶流形重排序的无监督特征选择算法（DMRR）的性能，我们在 8 个数据集上进行实验。这些数据集都是特征选择算法常用的数据集，具体细节如表 1 所列。

表 1 数据集描述

Table1 Dataset descriptions

| Datasets | Samples | Features | Classes |
|---|---|---|---|
| WARPAR10P | 130 | 2400 | 10 |
| YALE | 165 | 1024 | 15 |
| LUNG | 203 | 3312 | 5 |
| JAFFE | 213 | 676 | 10 |
| RELATHE | 1427 | 4322 | 2 |
| COIL20 | 1440 | 1024 | 20 |
| BASEHOCK | 1993 | 4862 | 2 |
| MADELON | 2600 | 500 | 2 |

### 4.2 实验对比算法

DMRR 作为一种后处理算法，可以对其他无监督特征选择算法输出的特征得分进行改进从而进一步增强性能。本文使用 3 种经典的无监督特征选择方法，即基于 LLE 的过滤式特征选择算法（LLEScore）[10]、基于拉普拉斯得分的特征选择算法（LapScore）[9]、多类簇特征选择算法（MCFS）[12]，与两种基于无监督特征选择的后处理算法 GRM[14] 和 AGRM[15] 组成以下 3 组对比算法。

第一组 LapScore[9]：对原始数据集，运用拉普拉斯特征选择算法进行特征选择。

LapScore_AGRM：对原始数据集，运用拉普拉斯特征选择算法得到特征评分，然后 AGRM 算法利用该特征评分信息进行特征选择。

LapScore_GRM：对原始数据集，运用拉普拉斯特征选择算法得到特征评分，然后 GRM 算法利用该特征评分信息进行特征选择。

LapScore_DMRR：对原始数据集，运用拉普拉斯特征选择算法得到特征评分，然后本文提出的基于对偶流形重排序的无监督特征选择算法利用该特征评分信息进行特征选择。

第二组 MCFS[12]：对原始数据集，运用多类簇特征选择算法进行特征选择。

MCFS_AGRM：对原始数据集，运用多类簇特征选择算法得到特征评分，然后 AGRM 算法利用该特征评分信息进行特征选择。

MCFS_GRM：对原始数据集，运用多类簇特征选择算法得到特征评分，然后 GRM 算法利用该特征评分信息进行特征选择。

MCFS_DMRR：对原始数据集，运用多类簇特征选择算法得到特征评分，然后本文提出的基于对偶流形重排序的无监督特征选择算法利用该特征评分信息进行特征选择。

第三组 LLEScore[10]：对原始数据集，运用 LLEScore 特征选择方法进行特征选择。

LLEScore_AGRM：对原始数据集，运用 LLE Score 特征选择算法得到特征评分，然后 AGRM 算法利用该特征评分信息进行特征选择。

LLESocre_GRM：对原始数据集，运用 LLEScore 特征选择算法得到特征评分，然后 GRM 算法利用该特征评分信息进行特征选择。

LLEScore_DMRR：对原始数据集，运用 LLEScore 特征选择算法得到特征评分，然后本文提出的基于对偶流形重排序的无监督特征选择算法利用该特征评分信息进行特征选择。

### 4.3 参数设置

对于参数的选择，本文在构造相似性矩阵时设置邻域的大小 $k=5$，高斯函数中参数 $\gamma=8$。对于模型中涉及的其他参数，$\lambda_1$ 从$\{10^0, 10^1, 10^2, 10^3, 10^4, 10^5\}$中选择，$\lambda_2$ 从$\{10^0, 10^1, 10^2, 10^3, 10^4, 10^5\}$中选择，此外评估算法性能时所选特征数量的集合为[10∶10∶100]。由于 $K$-means 算法聚类结果对初始化很敏感，因此本文运行 $K$-means 算法 20 次并报告其平均结果。

### 4.4 评价指标

本实验使用聚类准确性（Accuracy，ACC）、归一化互信息（Normalized Mutual Information，NMI）和聚类纯度（Purity）这 3 种聚类方法中常用的评价指标来评估算法的性能。指标的值越大，表示聚类性能越好。

聚类准确性（ACC）是用来表示所聚的类与样本原始类之间的一对一关系。给定一个样本点$\boldsymbol{x}_i$，$\boldsymbol{p}_i$ 和$\boldsymbol{q}_i$，分别用来表示聚类结果和样本的真实标签。则 $ACC$ 为：

$$ACC=\frac{1}{n}\sum_{i=1}^{n}\delta(\boldsymbol{q}_i, map(\boldsymbol{p}_i)) \tag{21}$$

其中，$n$ 表示样本数量，$\delta(x,y)$为判别函数，若 $x=y$，则 $\delta(x,y)$的值为 1，否则为 0。$map(\ )$ 是一个置换函数，用于将每一个簇索引映射到一个真实的标签中。

归一化互信息（NMI）主要用来表示聚类的质量。记 $\boldsymbol{C}$ 是真实类标签的集合，$\boldsymbol{C}'$是通过聚类算法计算的类标签集合，则它们的互信息 $MI(\boldsymbol{C},\boldsymbol{C}')$ 为：

$$MI(\boldsymbol{C},\boldsymbol{C}')=\sum_{\boldsymbol{c}_i\in \boldsymbol{C},\boldsymbol{c}'_j\in \boldsymbol{C}'} p(\boldsymbol{c}_i,\boldsymbol{c}_j{}')\log\frac{p(\boldsymbol{c}_i,\boldsymbol{c}_j{}')}{p(\boldsymbol{c}_i)p(\boldsymbol{c}_j{}')} \tag{22}$$

其中，$p(\boldsymbol{c}_i)$与 $p(\boldsymbol{c}_j{}')$分别是从数据集中任意选定一个样本点属于类$\boldsymbol{c}_i$ 和 $\boldsymbol{c}_j{}'$的概率，$p(\boldsymbol{c},\boldsymbol{c}_j{}')$是这个数据点同时属于这两个类的概率。因此归一化互信息为：

$$NMI(\boldsymbol{C},\boldsymbol{C}')=\frac{MI(\boldsymbol{C},\boldsymbol{C}')}{\max(H(\boldsymbol{C}),H(\boldsymbol{C}'))} \tag{23}$$

其中. $H(\boldsymbol{C})$和 $H(\boldsymbol{C}')$分别是 $\boldsymbol{C}$ 和 $\boldsymbol{C}'$的熵。

聚类纯度（Purity）主要是利用样本的正确率来对算法的聚类效果进行评价，公式如下：

$$Purity=\frac{1}{n}\sum_{k}\max(\boldsymbol{c}_k{}',\boldsymbol{c}_j) \tag{24}$$

其中，$n$ 代表样本个数，$\boldsymbol{C}=\{\boldsymbol{c}_1,\boldsymbol{c}_2,\cdots,\boldsymbol{c}_k\}$为真实样本中的类簇集合，$\boldsymbol{C}'=\{\boldsymbol{c}_1{}',\boldsymbol{c}_2{}',\cdots,\boldsymbol{c}_k{}'\}$为聚类后得到的分类。

### 4.5 实验结果分析

因为特征选择的最优数量是未知的，为了更好地评估无监督特征选择算法，我们对比了在多个数据集上选择不同特征数量的平均聚类性能，结果如表 2－表 4 所列，其中最优结果用加粗表示，次优结果用下划线标出。

表 2 不同特征选择算法在不同数据集上的聚类结果(ACC)

Table 2 Clustering results(ACC) of different feature selection algorithms on different datasets

| Algs\Ds | YALE | LUNG | JAFFE | RELATHE | COIL20 | BASEHOCK | MADELON | WARPAR1P | AVERAGE |
|---|---|---|---|---|---|---|---|---|---|
| LapScore | 0.3936 | 0.5489 | 0.7064 | 0.5364 | 0.5029 | 0.5029 | 0.5192 | 0.2970 | 0.5009 |
| LapScore_AGRM | 0.3982 | 0.5137 | 0.7353 | 0.5417 | 0.5380 | 0.5025 | **0.5742** | 0.3018 | 0.5132 |
| LapScore_GRM | 0.3938 | 0.4235 | 0.7208 | 0.5472 | 0.5409 | 0.5021 | 0.5267 | 0.3425 | 0.4997 |
| LapScore_DMRR | **0.4135** | **0.6678** | **0.7451** | **0.5506** | **0.5604** | **0.5422** | 0.5578 | **0.4601** | **0.5622** |
| MCFS | 0.3565 | 0.5478 | 0.6945 | 0.5405 | 0.5689 | 0.5075 | 0.5180 | 0.2333 | 0.4959 |
| MCFS_AGRM | 0.3514 | 0.5163 | 0.6909 | 0.5402 | 0.5712 | 0.5058 | 0.5176 | 0.2700 | 0.4954 |
| MCFS_GRM | 0.3443 | 0.3868 | 0.6754 | 0.5461 | 0.5715 | 0.5035 | 0.5167 | 0.3386 | 0.4854 |
| MCFS_DMRR | **0.4110** | **0.6716** | **0.7352** | **0.5506** | **0.5920** | **0.5394** | **0.5562** | **0.4670** | **0.5654** |
| LLEScore | 0.3210 | 0.5656 | 0.6884 | 0.5488 | **0.5870** | 0.5056 | 0.5211 | 0.3033 | 0.5051 |
| LLEScore_AGRM | 0.3400 | 0.3651 | 0.6322 | 0.5459 | 0.5556 | 0.5021 | 0.5087 | 0.3308 | 0.4725 |
| LLEScore_GRM | 0.3425 | 0.3435 | 0.7035 | 0.5464 | 0.5476 | 0.5020 | 0.5078 | 0.3411 | 0.4793 |
| LLEScore_DMRR | **0.4192** | **0.6621** | **0.7325** | **0.5511** | 0.5587 | **0.5410** | **0.5921** | **0.4582** | **0.5644** |

表 3 不同特征选择算法在不同数据集上的聚类结果(NMI)

Table 3 Clustering results(NMI) of different feature selection algorithms on different datasets

| Algs\Ds | YALE | LUNG | JAFFE | RELATHE | COIL20 | BASEHOCK | MADELON | WARPAR1P | AVERAGE |
|---|---|---|---|---|---|---|---|---|---|
| LapScore | 0.4456 | 0.4068 | 0.7884 | 0.0034 | 0.6286 | 0.0029 | 0.0041 | 0.2998 | 0.3225 |
| LapScore_AGRM | 0.4516 | 0.3103 | 0.7898 | 0.0032 | 0.6629 | 0.0027 | 0.0164 | 0.2926 | 0.3162 |
| LapScore_GRM | 0.4445 | 0.1806 | 0.7679 | 0.0020 | 0.6584 | 0.0019 | 0.0068 | 0.3367 | 0.2999 |
| LapScore_DMRR | **0.4706** | **0.4742** | **0.8098** | **0.0059** | **0.6852** | **0.0103** | **0.0173** | **0.5124** | **0.3732** |
| MCFS | 0.4111 | 0.3489 | 0.7467 | 0.0017 | 0.6919 | 0.0045 | 0.0015 | 0.1769 | 0.2979 |
| MCFS_AGRM | 0.4104 | 0.2946 | 0.7235 | 0.0017 | 0.6921 | 0.0037 | 0.0015 | 0.2382 | 0.2957 |
| MCFS_GRM | 0.4047 | 0.1303 | 0.7259 | 0.0021 | 0.6928 | 0.0008 | 0.0014 | 0.3368 | 0.2868 |
| MCFS_DMRR | **0.4708** | **0.4739** | **0.7824** | **0.0047** | **0.7060** | **0.0104** | **0.0141** | **0.5175** | **0.3725** |
| LLEScore | 0.3814 | 0.3805 | 0.7446 | **0.0092** | **0.7022** | 0.0049 | 0.0023 | 0.3099 | 0.3169 |
| LLEScore_AGRM | 0.3940 | 0.0917 | 0.6567 | 0.0010 | 0.6585 | 0.0010 | 0.0002 | 0.3057 | 0.2636 |
| LLEScore_GRM | 0.3950 | 0.0658 | 0.7493 | 0.0011 | 0.6317 | 0.0010 | 0.0002 | 0.3302 | 0.2718 |
| LLEScore_DMRR | **0.4833** | **0.4666** | **0.7788** | 0.0039 | 0.6814 | **0.0120** | **0.0272** | **0.5116** | **0.3706** |

表 4 不同特征选择算法在不同数据集上的聚类结果(Purity)

Table 4 Clustering results(Purity) of different feature selection algorithms on different datasets

| Algs\Ds | YALE | LUNG | JAFFE | RELATHE | COIL20 | BASEHOCK | MADELON | WARPAR1P | AVERAGE |
|---|---|---|---|---|---|---|---|---|---|
| LapScore | 0.4166 | 0.8262 | 0.7505 | 0.5473 | 0.5587 | 0.5029 | 0.5192 | 0.3076 | 0.5536 |
| LapScore_AGRM | 0.4175 | 0.7539 | 0.7712 | 0.5489 | 0.5765 | 0.5029 | **0.5742** | 0.3137 | 0.5574 |
| LapScore_GRM | 0.4124 | 0.7187 | 0.7537 | 0.5485 | 0.5765 | 0.5024 | 0.5267 | 0.3576 | 0.5496 |
| LapScore_DMRR | **0.4331** | **0.8582** | **0.7816** | **0.5517** | **0.5970** | **0.5422** | 0.5578 | **0.4969** | **0.6023** |
| MCFS | 0.3816 | 0.8010 | 0.7299 | 0.5468 | 0.6128 | 0.5075 | 0.5180 | 0.2451 | 0.5428 |
| MCFS_AGRM | 0.3702 | 0.7715 | 0.7185 | 0.5468 | 0.6094 | 0.5059 | 0.5176 | 0.2860 | 0.5407 |
| MCFS_GRM | 0.3647 | 0.6981 | 0.7107 | 0.5475 | 0.6089 | 0.5036 | 0.5167 | 0.3575 | 0.5385 |
| MCFS_DMRR | **0.4311** | **0.8471** | **0.7656** | **0.5514** | **0.6345** | **0.5394** | **0.5562** | **0.5047** | **0.6037** |
| LLEScore | 0.3457 | 0.8132 | 0.7269 | **0.5569** | **0.6227** | 0.5057 | 0.5211 | 0.3162 | 0.5511 |
| LLEScore_AGRM | 0.3616 | 0.6882 | 0.6627 | 0.5466 | 0.6070 | 0.5021 | 0.5087 | 0.3519 | 0.5286 |
| LLEScore_GRM | 0.3652 | 0.6866 | 0.7343 | 0.5468 | 0.5754 | 0.5021 | 0.5078 | 0.3605 | 0.5348 |
| LLEScore_DMRR | **0.4378** | **0.8508** | **0.7614** | 0.5512 | 0.5935 | **0.5411** | **0.5921** | **0.4971** | **0.6031** |

从表 2—表 4 可以得到以下结论：

(1)相较于其他方法，DMRR 模型在所有数据集上都取得了较好的聚类准确度。虽然其在 MADELON 和 COIL20 数据集上有一小部分没有取得最优结果，但也获得了次优的聚类准确率。并且在聚类指标 NMI 和 Purity 方面，本文模型的大部分结果都优于其他方法。

(2)原特征选择方法 LapScore，MCFS，LLEScore 经过 DMRR 模型处理后，在准确性 ACC 方面分别得到了不同程度的提升，分别为 12.24%，14.01%，11.74%；在互信息 NMI 与 Purity 方面也得到了不同程度的提升，分别为 15.72%，25.04%，16.95%，8.8%，11.22%，9.44%。这些数据直观地说明了，DMRR 模型作为一种后处理方法在提升原始特征选择算法性能方面是有效的，可以实现对其性能的进一步提升。

(3)通过与其他后处理算法 GRM 和 AGRM 的对比，可以看出 DMRR 模型在大多数数据集上是优于对比方法的。具体地，在聚类准确率 ACC 指标上，可以看到本文方法在所有数据集上的平均准确率是最高的，这充分体现了样本与特征之间对偶关系与样本的重要性信息在特征选择任务中是有着积极作用的。

图 3—图 8 给出了所有算法在数据集 YALE 和 WARPAR10P 上选择不同数量特征的聚类表现。从图中可以看出，在数据集 YALE 和 WARPAR10P 上，选择不同数量的特征用于聚类时，DMRR 模型所选择特征的聚类准确率 ACC、互信息 NMI、纯度 Purity 都明显高于其他方法。这充分说明样本重要性信息与样本特征之间的对偶关系对于特征选择任务来说是重要的，尤其在一些数据集上，这两种信息带来的性能提升更为明显。

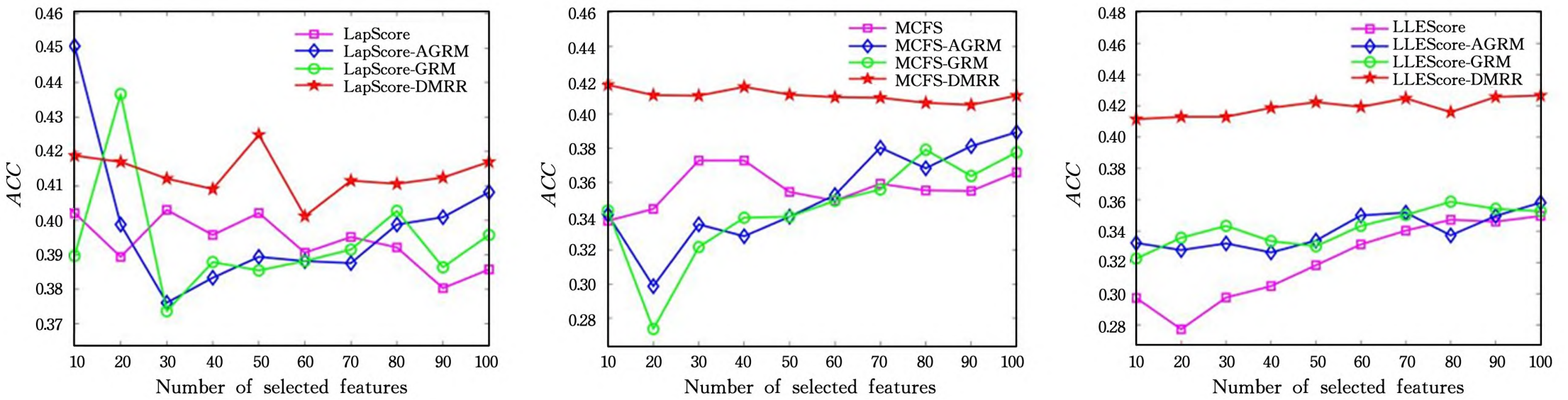

图 3　在 YALE 上选择不同数量特征对应的 *ACC*

Fig. 3　*ACC* of all methods with different number of selected features on YALE

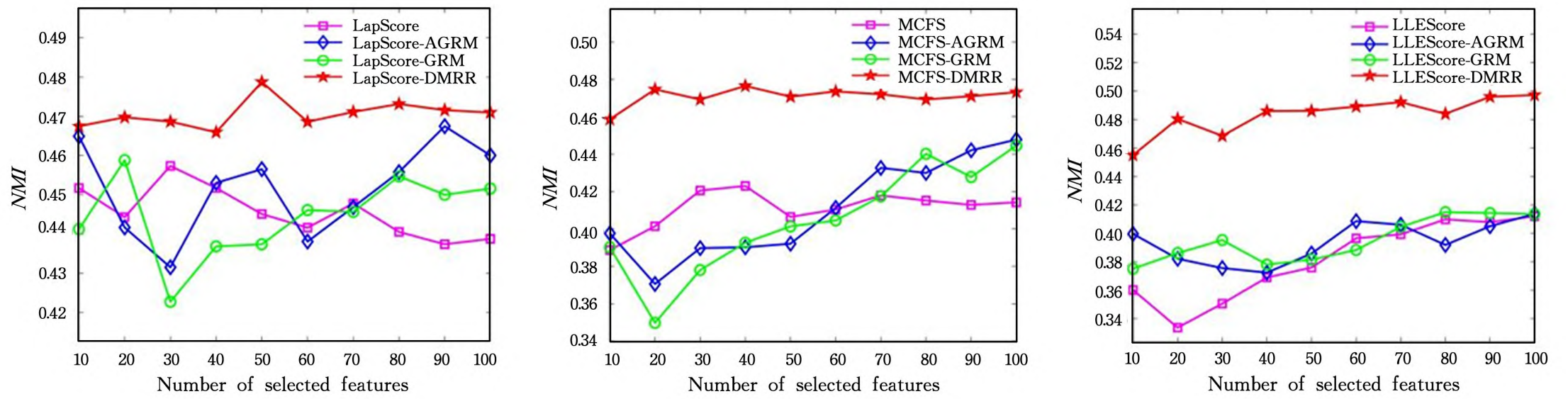

图 4　在 YALE 上选择不同特征数量对应的 *NMI*

Fig. 4　*NMI* of all methods with different number of selected features on YALE

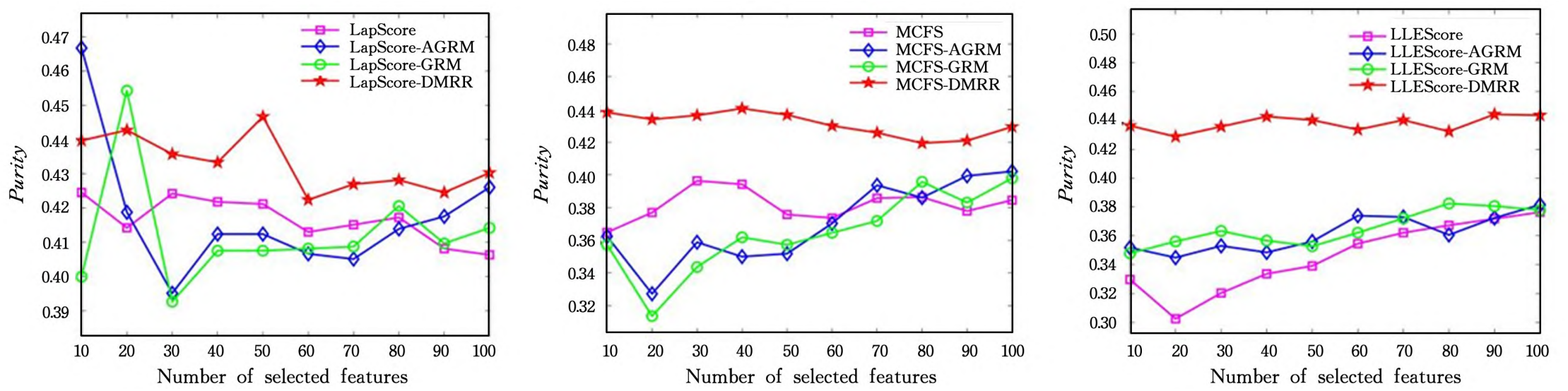

图 5　在 YALE 上选择不同特征数量对应的的 *Purity*

Fig. 5　*Purity* of all methods with different number of selected features on YALE

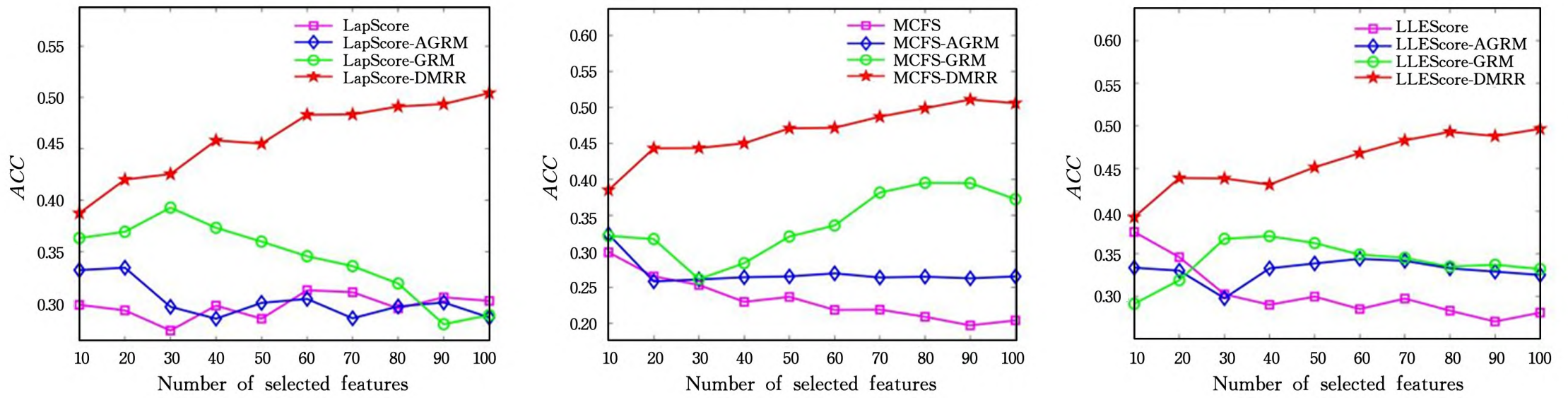

图 6　在 WARPAR10P 上选择不同特征数量对应的 *ACC*

Fig. 6　*ACC* of all methods with different number of selected features on WARPAR10P

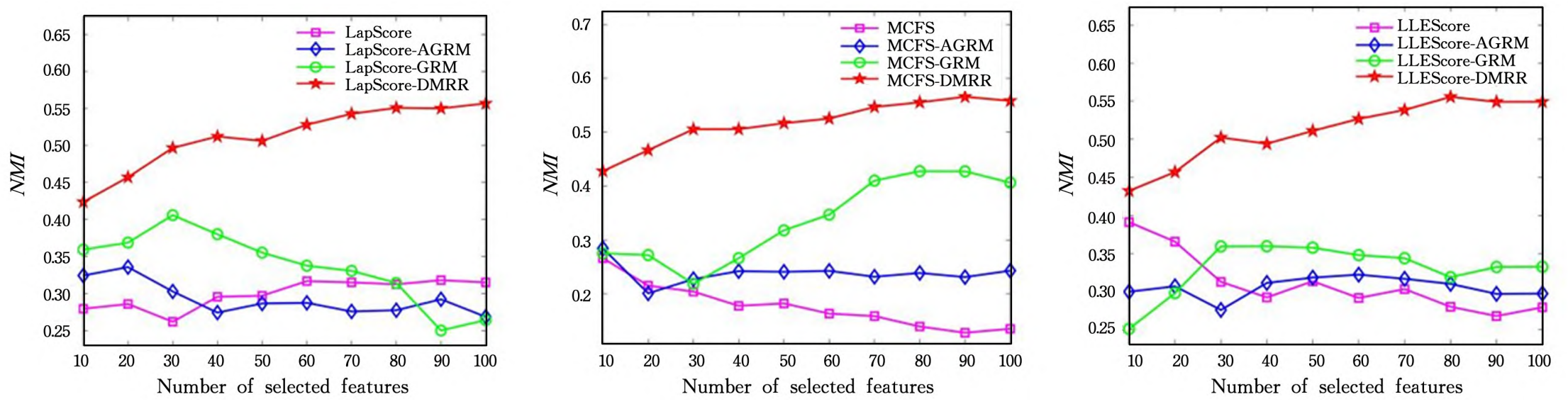

图 7　在 WARPAR10P 上选择不同特征数量对应的的 *NMI*

Fig. 7　*NMI* of all methods with different number of selected features on WARPAR10P

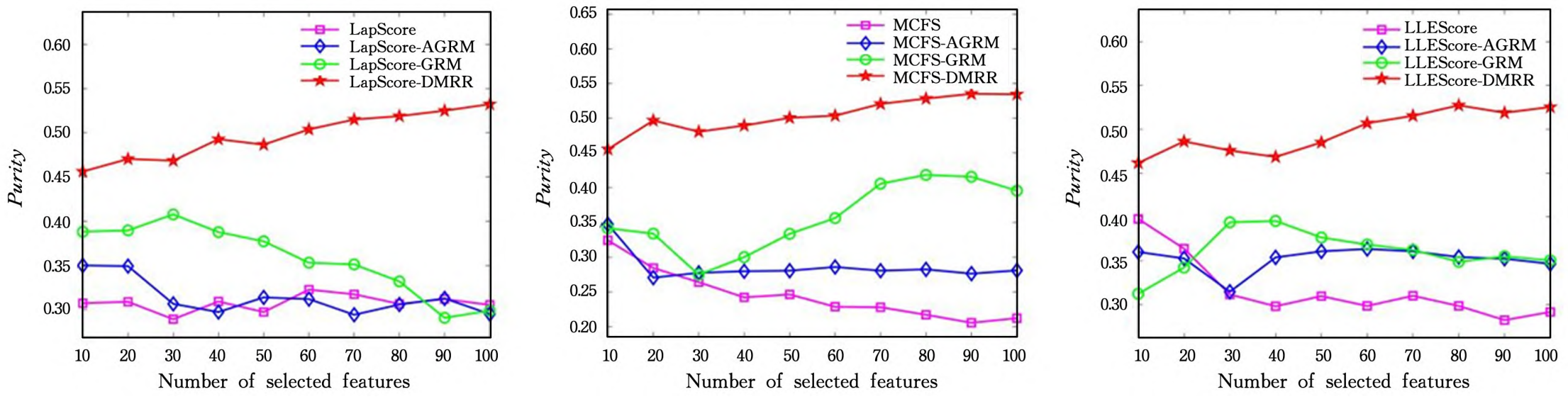

图 8 在 WARPAR10P 上选择不同特征数量对应的的 *Purity*

Fig. 8 *Purity* of all methods with different number of selected features on WARPAR10P

**结束语** 本文基于样本得分、特征得分、样本与特征之间的对偶关系，结合流形学习提出了一种基于对偶流形重排序的无监督特征选择算法。该算法是一种后处理算法，可以将特征选择算法输出的特征得分作为输入，联合样本与特征的对偶关系来对原始算法的性能进行改进。在多个数据集上的实验结果验证了本文算法在大多数情况下的聚类性能是优于原始特征选择方法的；并且在与两个后处理算法 GRM 和 AGRM 的比较中发现，本文算法在大多数情况下对原始特征选择算法性能的提升优于 GRM 与 AGRM 算法，这在一定程度上说明了本文算法的有效性。更进一步地，从实验中可以得出结论：不同样本的重要性信息、样本与特征之间的对偶关系对于特征选择任务是重要的。未来我们将进一步结合样本与特征之间的对偶关系来进行特征选择任务的探索。

## 参 考 文 献

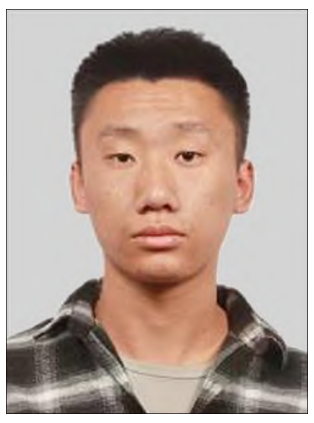

**LIANG Yunhui**，born in 1998，postgraduate，is a member of China Computer Federation. His main research interests include data mining and feature selection.

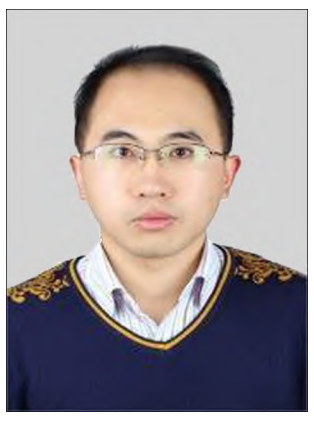

**DU Liang**，born in 1985，Ph.D，associate professor，is a member of China Computer Federation. His main research interests include data mining，feature selection and clustering analysis.

(责任编辑：何杨)